\documentclass[conference]{IEEEtran}
\IEEEoverridecommandlockouts
\usepackage{amsmath,amssymb,amsfonts}
\usepackage{graphicx}
\usepackage{textcomp}
\usepackage{xcolor}
\usepackage{bm}
\usepackage{amsmath}
\usepackage{amssymb}
\usepackage{amsthm}
\usepackage{mathrsfs}
\usepackage{enumerate}
\usepackage{multirow}
\usepackage{color}
\usepackage{threeparttable}
\usepackage{subfigure}

\usepackage{booktabs}
\usepackage[table]{xcolor}
\usepackage{adjustbox}
\usepackage{bbold}
\newcommand{\gain}[1]{{\footnotesize\color{green!50!black}(#1)}}
\usepackage{wrapfig}
\usepackage{tabularx}
\usepackage{makecell}

\usepackage[square, comma, sort&compress, numbers]{natbib}
\usepackage[pagebackref=false,breaklinks=true,letterpaper=true,colorlinks,bookmarks=false]{hyperref}
\usepackage{times}
\usepackage{breakurl}
\usepackage{array}
\usepackage{verbatim}
\usepackage{algorithm}
\usepackage{bbding}
\usepackage{algpseudocode}
\usepackage{lettrine}

\def\BibTeX{{\rm B\kern-.05em{\sc i\kern-.025em b}\kern-.08em
    T\kern-.1667em\lower.7ex\hbox{E}\kern-.125emX}}
\begin{document}

\title{Reading Decoder Trajectories: Training-Free Counterfactual Query-Trajectory Reliability for Small-Object Detection}

\author{
\IEEEauthorblockN{
Zhaoning Shi and Bo Ma
}
\IEEEauthorblockA{
\textit{Beijing Institute of Technology}\\
Beijing, China\\
\{3220235133, bma000\}@bit.edu.cn
}
}

\maketitle

\begin{abstract}
Small-object detection remains challenging because limited pixels cause information loss and suppress the scale knowledge encoded in pretrained detectors. Existing approaches mainly improve representations through multiscale training, architecture redesign, or parameter adaptation, implicitly assuming that frozen models lack the required capability. We challenge this assumption and hypothesize that small-object knowledge already exists in frozen detectors but remains underactivated and unstable during query evolution. To test this hypothesis, we propose Counterfactual Query-Trajectory Reliability (CQTR), a training-free framework that elicits latent responses through counterfactual scale interventions and interprets candidate reliability from decoder-internal spatial convergence, semantic persistence, and cross-scale conflicts. A small unlabeled training subset selects the appropriate correction mechanism for each model–data stream, without parameter updates or target-domain annotations. Across 27 combinations of nine frozen detectors and three datasets, CQTR consistently improves average precision (AP) and average precision for small objects (\(\mathrm{AP}_{s}\)). Closed-loop analyses further show that scale intervention activates latent responses, trajectory evidence predicts ground-truth support, and unlabeled routing selects the more effective branch. CQTR therefore reframes small-object detection from external scale augmentation to the activation and reliability assessment of latent scale knowledge.

\end{abstract}


\section{Introduction}\label{sec:introduction}

Small-object detection is fundamental to visual perception systems such as autonomous driving, unmanned aerial vehicle perception, and remote-sensing analysis. Because a small object occupies only a few pixels and feature cells, its visual evidence is easily weakened by spatial downsampling and overwhelmed by background responses. Detecting such objects therefore requires not only sufficient feature resolution, but also the reliable formation of object hypotheses from limited internal evidence. This raises a fundamental question: must small-object capability be acquired through additional training, or is the required scale knowledge already encoded in a pretrained detector but insufficiently activated during standard inference?

Set-prediction detectors represented by the Detection Transformer (DETR) formulate object detection through learnable queries and global matching \citep{carion2020detr}. Subsequent studies improve their query mechanisms through deformable sampling, dynamic anchors, query selection, and efficient encoding \citep{zhu2021deformable,liu2022dab,zhang2023dino,huang2025deim}. Nevertheless, their performance on small objects remains substantially lower than that on larger objects. Existing approaches commonly address this gap through multiscale training, high-resolution features, image slicing, or parameter adaptation. These methods improve external representations but do not determine whether the resulting gains come from newly learned knowledge or the reactivation of capabilities already present in a frozen model. Meanwhile, existing interpretability methods mainly attribute final predictions to input regions or attention distributions. They provide limited insight into how queries progressively establish spatial and semantic decisions during decoding.

\begin{figure*}[t]
\centering
\includegraphics[width=\linewidth]{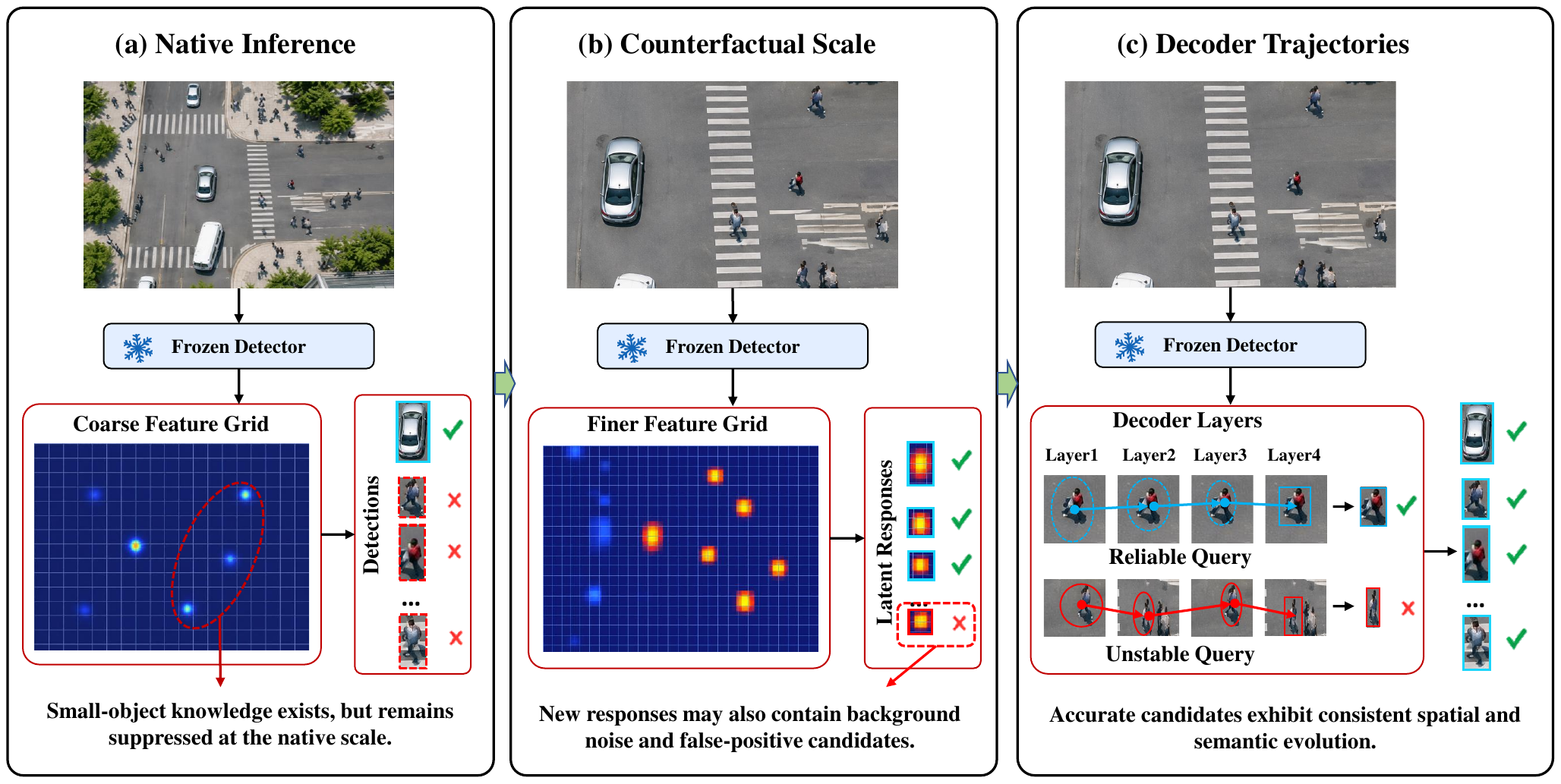}
\vspace{-8mm}
\caption{Motivation for CQTR. Counterfactual scaling activates latent small-object evidence in frozen detectors, while spatial-semantic decoder trajectories distinguish reliable queries from amplified noise.}
\vspace{-4mm}
\label{fig:motivation}
\end{figure*}

We argue that small-object knowledge does not completely disappear from frozen detectors. Instead, spatial compression may leave this knowledge latent or unstable during query evolution. Increasing the inference scale provides a counterfactual observation in which compressed object evidence can be reintroduced into the decoding process. However, the same intervention may also amplify background textures and spurious candidates. The central problem is therefore not simply to combine predictions from multiple scales, but to identify which scale-activated responses reflect reliable object knowledge by examining their internal query dynamics, as illustrated in Fig.~\ref{fig:motivation}.

To address this problem, we propose Counterfactual Query-Trajectory Reliability (CQTR), a training-free framework for activating and assessing latent scale knowledge in frozen detectors. CQTR constructs a counterfactual scale branch and interprets intermediate decoder states through spatial convergence and semantic persistence. It further detects unreliable candidates from semantic conflicts between native-scale and counterfactual predictions. Because query trajectories and cross-scale conflicts characterize different failure modes, CQTR uses a small unlabeled subset of the corresponding training split to select the appropriate correction mechanism for each model-data stream. The entire framework keeps the detector architecture and parameters unchanged, requires no backpropagation, and accesses no target-domain training annotations.

We evaluate CQTR across 27 combinations of nine frozen detectors and three datasets, including COCO2017, TinyPerson, and VisDrone2019. CQTR consistently improves AP, while \(\mathrm{AP}_{s}\) increases in every setting. Closed-loop analyses show that counterfactual scaling activates latent small-object responses, query-trajectory evidence predicts candidate reliability, and unlabeled routing selects the more effective correction mechanism. Internal heatmaps and detection visualizations further connect changes in decoder states to improved small-object recall. These results support the view that small-object detection can be approached as the activation and reliability assessment of latent scale knowledge rather than solely as external scale augmentation.
The main contributions are as follows:

\begin{itemize}
\item We reformulate small-object failure in frozen detectors as a problem of latent scale-knowledge activation and reliability assessment, distinguishing underactivated knowledge from genuinely missing capability.
\item We introduce CQTR, a training-free framework that interprets and corrects scale-activated candidates through counterfactual intervention, spatial-semantic query trajectories, and cross-scale semantic conflicts.
\item We conduct extensive cross-model and cross-domain experiments. Consistent improvements, closed-loop analyses, and internal visualizations establish an evidence chain from latent knowledge activation to query assessment and improved small-object detection.
\end{itemize}

\section{Related Work}

\subsection{Small-Object Detection}\label{sec:related-small}

Existing small-object detectors can be grouped into representation enhancement, spatial decomposition, and query optimization. Representation-enhancement methods use feature pyramids, high-resolution branches, or super-resolution reconstruction to recover fine-grained information ~\citep{lin2017fpn,yang2022querydet}. Spatial-decomposition methods increase relative object resolution through image slicing, local magnification, or density-guided cropping ~\citep{singh2018snip,singh2018sniper,najibi2019autofocus,akyon2022sahi,kisantal2019augmentation}. Query-optimization methods improve sampling locations, reference points, and matching mechanisms to strengthen small-object modeling in set-prediction detectors ~\citep{carion2020detr,zhu2021deformable,liu2022dab,zhang2023dino,huang2025deim}. These approaches primarily attribute the bottleneck to insufficient visible evidence and learn new scale representations through architecture design or parameter optimization. They do not determine whether a frozen detector has already encoded the relevant knowledge. In contrast, we add neither feature levels nor retraining. We expose latent small-object responses through counterfactual scale intervention and study their reliability during decoding.

\begin{figure*}[t]
\centering
\includegraphics[width=\linewidth]{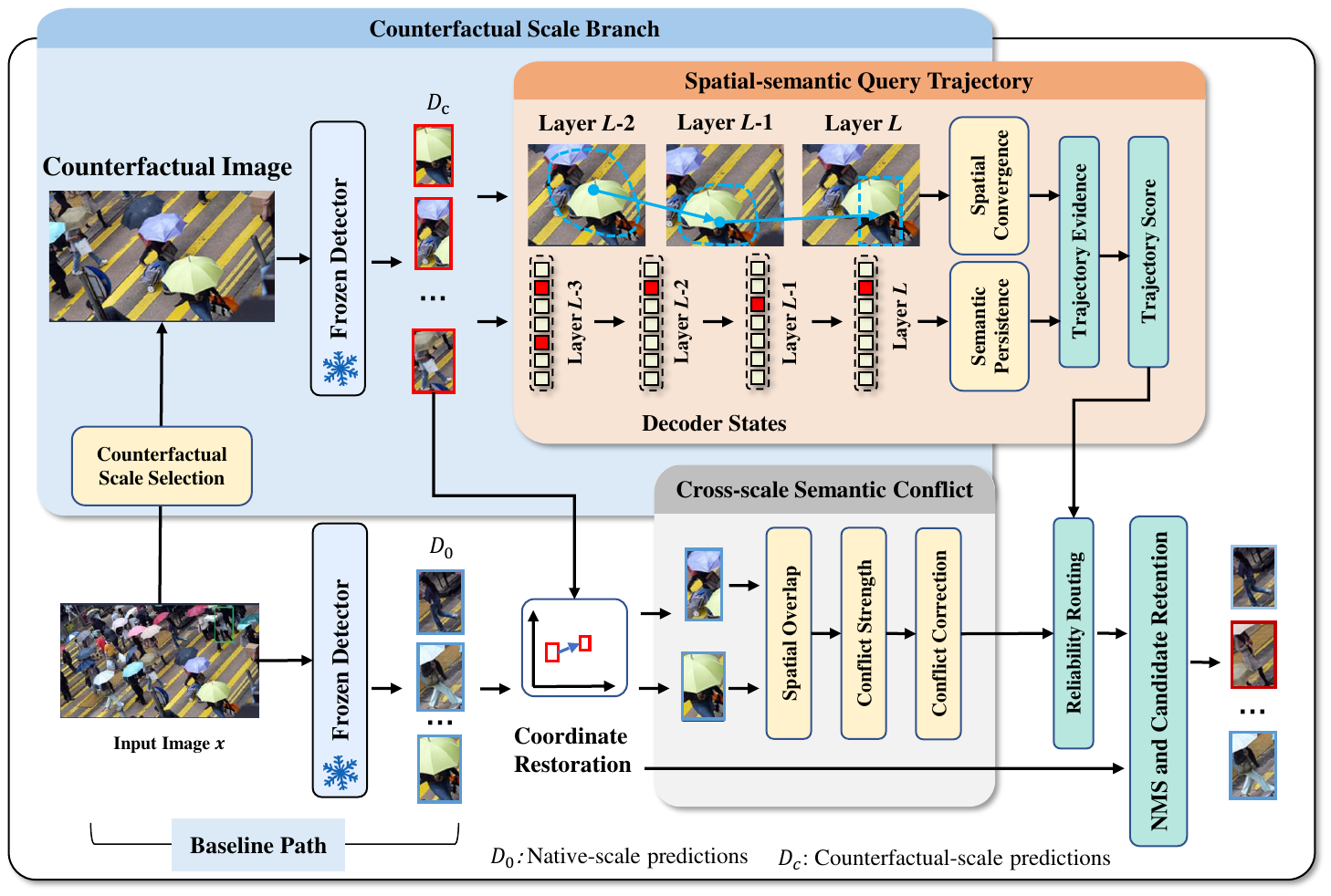}
\vspace{-8mm}
\caption{Overview of CQTR. CQTR interprets native and counterfactual predictions through decoder trajectories and cross-scale semantic conflicts, then uses unlabeled reliability routing to select the correction stream for final fusion. Snowflakes denote frozen parameters.}
\vspace{-4mm}
\label{fig:overview}
\end{figure*}

\subsection{Internal Model Interpretation}\label{sec:related-interpretation}

Existing model-interpretation methods mainly comprise post hoc attribution and internal-mechanism analysis. Post hoc methods localize image regions that influence a final prediction through attention, saliency, or gradient responses ~\citep{selvaraju2017gradcam,chefer2021generic}. They reveal what the model attends to, but provide limited evidence of how a prediction forms inside the network. Internal-mechanism methods analyze intermediate features, neuron activations, or computational paths to identify the internal units and evolution associated with particular behaviors ~\citep{dosovitskiy2021vit,akyon2022sahi}. Most detection explanations, however, remain centered on final outputs and rarely examine how scale changes affect the positional convergence and class evolution of object queries. Rather than attributing predictions externally, CQTR reads decoder trajectories under a counterfactual scale intervention and jointly interprets candidate reliability through spatial convergence, semantic persistence, and cross-scale conflict. This moves small-object explanation from output visualization toward an analysis of query-formation mechanisms.

\section{Method}

\subsection{Problem Formulation and Overall Architecture}\label{sec:overview}

Given an input image $\bm{x}$ and a detector ${F}_{\theta }$ with frozen parameters, let ${S}_{0}$ denote the native preprocessing scale. The native branch produces ${D}_{0}$, where each candidate contains a bounding box, class label, and confidence score in the original-image coordinate system:
$D_0 = F_{\theta}\!\left(T_{S_0}(\bm{x})\right).$
Here, ${T}_{{S}_{0}}$ denotes the preprocessing transformation applied to the input image at native scale ${S}_{0}$. CQTR changes only the input scale to construct a counterfactual branch and retains box and class responses from the last several decoder layers. Trajectory interpretation and cross-scale conflict interpretation then correct the counterfactual candidates ${D}_{\mathrm{c}}$ in parallel, after which unlabeled routing selects the more reliable correction stream. We summarize the overall process as
$\widehat{D}=\mathcal{F}(D_0,D_c).$
Here, $\mathcal{F}(\cdot )$ denotes the training-free correction process composed of query-trajectory interpretation, cross-scale semantic-conflict interpretation, unlabeled stream routing, and candidate fusion, as shown in Fig.~\ref{fig:overview}.

\subsection{Counterfactual Scale Branch}\label{sec:counterfactual}

Native-scale resizing may compress small objects into only a few feature cells, while uniformly increasing the scale can amplify interpolation redundancy and background noise. CQTR therefore selects the counterfactual scale adaptively according to the input’s spatial compression relative to the native scale.

Given the height ${H}_{x}$ and width ${W}_{x}$ of the original image, we define the spatial compression ratio as
$
\rho(\bm{x})=\frac{\sqrt{H_xW_x}}{S_0}.
$
A larger ratio indicates stronger information compression under native resizing. The counterfactual scale is selected as
\begin{equation}
S(\bm{x})=\begin{cases}S_1, & \rho(\bm{x})\leq\tau_p,\\ S_2, & \rho(\bm{x})>\tau_p,\end{cases}
\end{equation}
where ${S}_{1}$ and ${S}_{2}$ are the lower and higher counterfactual scales, respectively, and ${\tau }_{p}$ is the spatial-compression threshold. With detector parameters frozen, the counterfactual forward pass is
$D_c=F_{\theta}\!\left(T_{S(\bm{x})}(\bm{x})\right).$
For a detector that dynamically derives spatial relations from the current feature shapes, we use its native forward pass directly. Otherwise, we save and temporarily clear the preset encoder and decoder inference sizes before this forward pass. This prompts the model to reconstruct positional embeddings, decoder anchors, and validity masks from the actual multiscale feature shapes. The original sizes are restored immediately afterward. Following spatial-semantic query-trajectory interpretation, counterfactual predictions are mapped back to the original-image coordinate system while the intermediate responses from the final decoder layers are retained.

\subsection{Three-Step Spatial-Semantic Query-Trajectory Interpretation}\label{sec:trajectory}

Final confidence ignores how a prediction evolves during decoding. CQTR therefore estimates reliability from spatial convergence over the last three box transitions and semantic persistence over the final four decoder layers, then continuously recalibrates counterfactual scores.

Assume that the counterfactual branch contains $L$ decoder layers. For final candidate $i$, let ${\mathit{b}}_{i}^{\left(l\right)}$ denote its normalized box at layer $l$, and let ${p}_{i,y}^{\left(l\right)}$ denote its class response. Its final predicted class is
$y_i=\arg\max_y p_{i,y}^{(L)}.$
First, we measure spatial convergence by the geometric mean of intersection over union (IoU) over the last three box transitions:
\begin{equation}
a_i=\exp\!\left[\frac{1}{3}\sum_{l=L-2}^{L}\log\!\left(\max\!\left(\operatorname{IoU}(\bm{b}_i^{(l-1)},\bm{b}_i^{(l)}),\epsilon\right)\right)\right].
\end{equation}
Here, $\epsilon$ is a small constant that prevents numerical instability. A larger ${a}_{i}$ indicates that the query box stabilizes toward the end of decoding, while a smaller ${a}_{i}$ indicates persistent positional jumps.

Spatial convergence alone does not guarantee class reliability. To measure persistent support for the final class throughout decoding, we compare the final-class response with the strongest class response in each of the last four layers:
\begin{equation}
c_i=\exp\!\left[\frac{1}{4}\sum_{l=L-3}^{L}\log\!\left(\max\!\left(\frac{p_{i,y_i}^{(l)}}{\max_y p_{i,y}^{(l)}},\epsilon\right)\right)\right].
\end{equation}
When the final class remains close to the strongest class in consecutive decoder layers, ${c}_{i}$ approaches 1. Frequent switching between classes reduces ${c}_{i}$. We then define query-trajectory evidence as the geometric mean of spatial convergence and semantic persistence:
$e_i=\sqrt{a_ic_i}.$
To avoid removing small objects with a fixed threshold, we recalibrate counterfactual confidence only relative to the median evidence of the current image:
\begin{equation}
s_i^{\mathrm{traj}}=\operatorname{clip}_{[0,1]}\!\left(s_iw_i^{\mathrm{traj}}\right),
\end{equation}
where
\begin{equation}
w_i^{\mathrm{traj}}=\operatorname{clip}_{[w_{\min},w_{\max}]}\!\left(\exp\!\left[\beta\left(e_i-\operatorname{med}(\bm{e})\right)\right]\right).
\end{equation}
Here, ${s}_{i}$ is the original confidence of counterfactual candidate i, $\bm{e}$ is the set of trajectory-evidence values for all counterfactual candidates in the current image, and $\beta$ controls the correction strength. The operator ${\operatorname{clip}}_{\left(a,b\right)}(\cdot )$ clips its input to the interval $\left(a,b\right)$.

\subsection{Cross-Scale Semantic-Conflict Interpretation}\label{sec:conflict}

Stable trajectories may still produce inconsistent semantic explanations across scales because of context drift or class competition. To preserve newly exposed objects, CQTR soft-suppresses only counterfactual candidates that strongly overlap high-confidence native predictions but predict different classes.

For counterfactual candidate $i$, let ${\bm{b}}_{i}^{c}$, ${y}_{i}^{c}$, and ${s}_{i}^{c}$ denote its box, class, and original confidence. The corresponding quantities for native candidate $k$ are ${\bm{b}}_{k}^{0}$, ${y}_{k}^{0}$, and ${s}_{k}^{0}$. After both prediction streams are restored to the original-image coordinate system, we define confidence-weighted cross-scale spatial overlap as
$w_{ik}=\operatorname{IoU}(\bm{b}_i^c,\bm{b}_k^0)\sqrt{s_k^0}.$
The maximum overlap with a native candidate of a different class defines the conflict strength:
$r_i=\max_{k:\,y_k^0\neq y_i^c}w_{ik}.$
A larger conflict strength indicates a stronger contradiction between the two scale-specific interpretations of the same region. The conflict-corrected score is
\begin{equation}
s_i^{\mathrm{conf}}=s_i^c\max\!\left(w_{\min},1-\gamma r_i\right),
\end{equation}
where $\gamma$ controls the suppression strength and ${w}_{\mathrm{m}\mathrm{i}\mathrm{n}}$ is the minimum retention factor.

Trajectory recalibration and conflict suppression form parallel candidate streams and are not applied sequentially. The unlabeled router in the next subsection determines which stream is used.

\begin{figure*}[t]
\centering
\includegraphics[width=\linewidth]{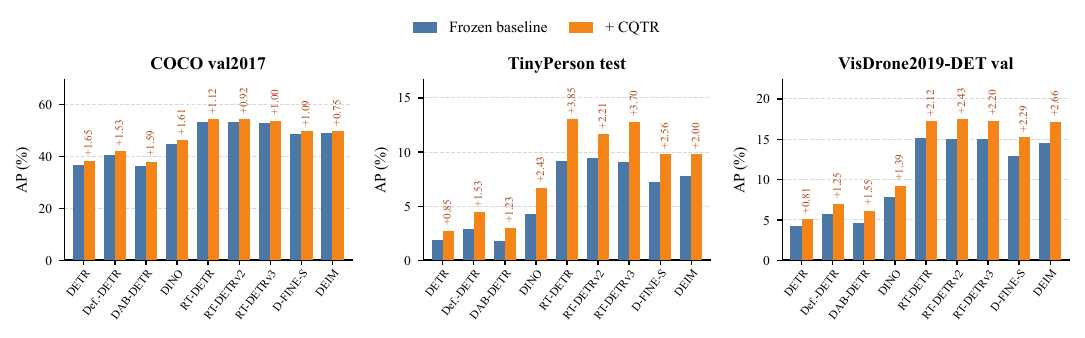}
\vspace{-8mm}
\caption{Overall AP comparison on COCO, TinyPerson, and VisDrone2019. Numbers above the orange bars denote absolute AP gains.}
\vspace{-3mm}
\label{fig:main_ap}
\end{figure*}

\begin{figure*}[t]
\centering
\includegraphics[width=\linewidth]{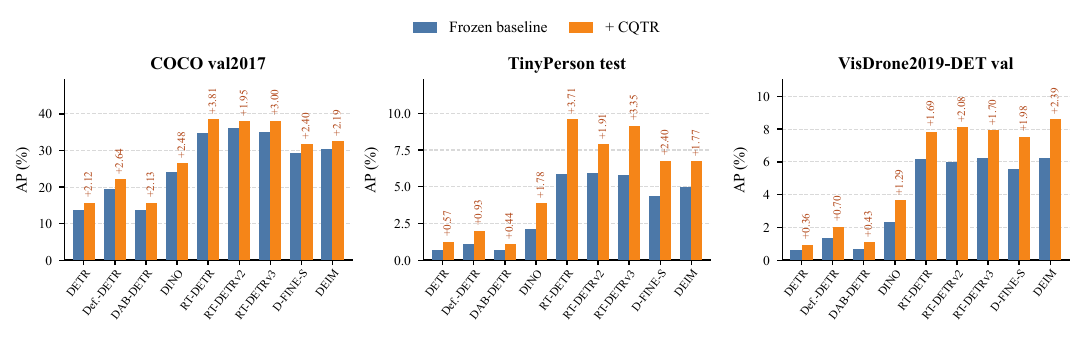}
\vspace{-8mm}
\caption{Small-object AP comparison on COCO, TinyPerson, and VisDrone2019. Numbers above the orange bars denote absolute improvements over the corresponding frozen baselines.}
\vspace{-4mm}
\label{fig:main_aps}
\end{figure*}

\begin{table*}[t]
\centering
\small
\caption{Component ablation of RT-DETR-R50 across the three datasets.}
\label{tab:ablation}
\resizebox{\textwidth}{!}{%
\begin{tabular}{lcccccc}
\toprule
Configuration & COCO AP & COCO $\mathrm{AP}_\mathrm{s}$ & Tiny AP & Tiny $\mathrm{AP}_\mathrm{s}$ & Vis AP & Vis $\mathrm{AP}_\mathrm{s}$ \\
\midrule
Frozen baseline & 53.09 & 34.74 & 9.19 & 5.89 & 15.11 & 6.16 \\
Counterfactual scale only & 54.18 & 38.51 & 12.99 & 9.41 & 17.13 & 7.88 \\
+ Trajectory recalibration & 54.21 & 38.55 & 12.96 & 9.39 & 17.12 & 7.86 \\
+ Conflict suppression & 54.16 & 38.44 & 13.04 & 9.60 & 17.23 & 7.85 \\
\rowcolor{blue!6} CQTR stream routing & 54.21 & 38.55 & 13.04 & 9.60 & 17.23 & 7.85 \\
\bottomrule
\end{tabular}%
}
\vspace{-3mm}
\end{table*}

\subsection{Unlabeled Stream-Level Reliability Routing and Unified Output}\label{sec:routing}

Query trajectories and cross-scale semantic conflicts capture complementary forms of internal instability, and applying both corrections indiscriminately may suppress useful signals. CQTR therefore estimates reliability from a small unlabeled sample and selects trajectory recalibration or conflict suppression for the entire model-data stream.

For each model-data stream, we sample $M$ unlabeled images from the training split using a fixed random seed. We first define stream-level trajectory dispersion as
$
\bar d=\frac{1}{M}\sum_{m=1}^{M}\operatorname{Std}_i\!\left(e_i^{(m)}\right).
$
Higher trajectory dispersion means that trajectory evidence more clearly separates candidate reliability. We define the conflict statistic as
$
\bar r=\frac{1}{M}\sum_{m=1}^{M}\max_i r_i^{(m)}.
$
A larger conflict statistic indicates stronger class competition caused by scale changes. Based on these statistics, we define the stream-level routing variable as
$
z=\mathbb{I}\!\left[\bar d\geq\tau_d\ \land\ \bar r\geq\tau_c\right],
$
where ${\tau }_{d}$ and ${\tau }_{c}$ are the thresholds for trajectory dispersion and conflict strength. The operator $1[ \cdot ]$ is the indicator function. We apply query-trajectory recalibration when $z=1$. Otherwise, we apply cross-scale semantic-conflict suppression. After unlabeled calibration, the routing variable remains fixed for the entire model-data stream. The corrected counterfactual predictions are
\begin{equation}
\widetilde{D}_c=\begin{cases}\{(\bm{b}_i^c,y_i^c,s_i^{\mathrm{traj}})\}_{i=1}^{Q}, & z=1,\\ \{(\bm{b}_i^c,y_i^c,s_i^{\mathrm{conf}})\}_{i=1}^{Q}, & z=0.\end{cases}
\end{equation}
Finally, the spatial compression ratio $\rho \left(\mathit{x}\right)$ defined in Section~\ref{sec:counterfactual} determines whether the native branch is retained:
\begin{equation}
\widehat{D}=\begin{cases}\operatorname{TopQ}\!\left(\operatorname{ClassNMS}(D_0\cup\widetilde{D}_c,\eta)\right), & \rho(\bm{x})\leq\tau_r,\\ \widetilde{D}_c, & \rho(\bm{x})>\tau_r,\end{cases}
\end{equation}
where $Q$ is the detector's candidate budget and $\eta$ is the threshold for class-aware non-maximum suppression (NMS).

\begin{figure*}[t]
\centering
\includegraphics[width=\linewidth]{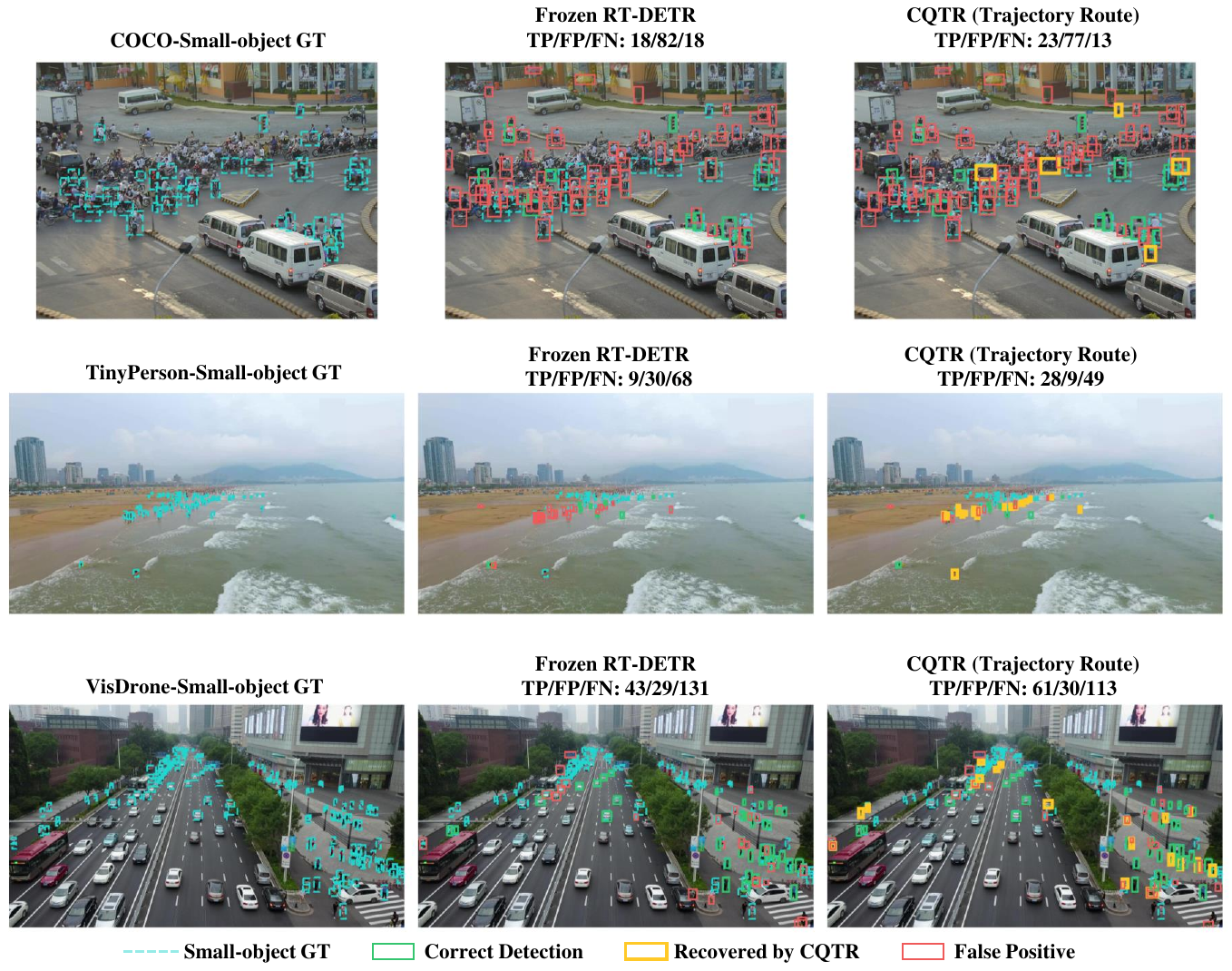}
\vspace{-6mm}
\caption{Final small-object detections produced by CQTR. Automatically selected examples at a shared confidence threshold show that CQTR recovers missed objects across all three datasets and reduces false positives on COCO and TinyPerson.}
\vspace{-3mm}
\label{fig:qualitative}
\end{figure*}

\section{Experiments}
\label{gen_inst}

\textbf{Datasets.} We evaluate on COCO val2017 \citep{lin2014coco}, TinyPerson test \citep{yu2020scalematch}, and VisDrone2019-DET val \citep{du2019visdrone}. COCO contains 5,000 validation images. TinyPerson and VisDrone contain 786 and 548 evaluation images, respectively. For the latter two datasets, we directly apply COCO-pretrained weights without accessing target-domain training annotations or performing fine-tuning, thereby testing generalization to cross-domain small-object scenes.

\textbf{Evaluation metrics.} All three datasets use the same COCO-style bounding-box evaluator. We report AP averaged over IoU thresholds, AP at IoU thresholds of 0.50 and 0.75 ($\mathrm{AP}_{50}$ and $\mathrm{AP}_{75}$), and $\mathrm{AP}_\mathrm{s}$. TinyPerson results do not follow its official miss-rate protocol. VisDrone results are also subject to the COCO-style limit of 100 maximum detections per image (maxDets=100). We therefore compare CQTR only with its paired frozen baseline under the same protocol and do not directly compare absolute values with published results obtained under different protocols.

\textbf{Implementation details.} We evaluate DETR-R50 \citep{carion2020detr}, Deformable DETR-R50 \citep{zhu2021deformable}, DAB-DETR-R50 \citep{liu2022dab}, DINO-R50 \citep{zhang2023dino}, RT-DETR-R50 \citep{zhao2024rtdetr}, RT-DETRv2-R50 \citep{lv2024rtdetrv2}, RT-DETRv3-R50 \citep{wang2025rtdetrv3}, D-FINE-S \citep{peng2025dfine}, and DEIM-D-FINE-S \citep{huang2025deim}. These models cover fixed queries, dynamic anchors, deformable sampling, query selection, and real-time hybrid encoding. Every model uses COCO-pretrained weights and runs with frozen parameters under inference mode without gradients. TinyPerson and VisDrone experiments use no target-domain training annotations and perform no fine-tuning. We use S0/S1/S2=640/800/960, ${\tau }_{p}=2.0$, ${\tau }_{r}=1.0$, an NMS threshold of 0.8, and at most 300 internal candidates. The trajectory interpretation uses the last three box transitions and the last four class responses with $\beta =0.05$. The semantic-conflict coefficient is $\mathrm{\gamma }=0.10$. For each model-dataset combination, we use random seed 31415 to sample 100 unlabeled images from the corresponding training split for one-time stream calibration, with ${\tau }_{d}=0.22$ and ${\tau }_{c}=0.85$. On COCO, calibration routes RT-DETR-R50, RT-DETRv3-R50, and DEIM-D-FINE-S to query-trajectory recalibration and the remaining models to cross-scale semantic-conflict suppression. Every model on TinyPerson and VisDrone is routed to the conflict branch. The selected route remains fixed throughout full evaluation. We use the same configuration for all nine detectors and three datasets. Full evaluation covers 5,000, 786, and 548 images, respectively, for a total of 27 model-dataset combinations. All experiments run on one NVIDIA RTX 3060 GPU.

\subsection{Results Across Detectors and Datasets}\label{sec:comparisons}

Fig.~\ref{fig:main_ap} and~\ref{fig:main_aps} compare CQTR with nine frozen detectors on COCO, TinyPerson, and VisDrone2019. The figures visualize overall AP and $\mathrm{AP}_\mathrm{s}$, respectively. Complete numerical results for AP, $\mathrm{AP}_{50}$, $\mathrm{AP}_{75}$, and \(\mathrm{AP}_\mathrm{s}\) are reported in Tables~\ref{tab:coco}--\ref{tab:visdrone} of Appendix~\ref{app:full_results}.

As shown in Figs.~\ref{fig:main_ap} and~\ref{fig:main_aps} and detailed in Tables~\ref{tab:coco}--\ref{tab:visdrone}, CQTR improves both AP and $\mathrm{AP}_{s}$ across all 27 detector-dataset combinations. On COCO, the gains reach 0.75--1.65 AP and 1.95--3.81 $\mathrm{AP}_{s}$, with RT-DETR-R50 improving from 53.09 to 54.21 AP and from 34.74 to 38.55 $\mathrm{AP}_{s}$. On TinyPerson, CQTR yields gains of 0.85--3.85 AP and 0.44--3.71 $\mathrm{AP}_{s}$, where RT-DETR-R50 increases from 9.19 to 13.04 AP. On VisDrone2019, the improvements range from 0.81 to 2.66 AP and from 0.36 to 2.39 $\mathrm{AP}_{s}$. DEIM-D-FINE-S achieves the largest AP gain, rising from 14.50 to 17.16, while RT-DETRv2-R50 obtains the highest absolute AP of 17.49. These consistent improvements across diverse detector families and target domains demonstrate that CQTR rec is neither architecture architecture-specific nor limited to the COCO scale distribution.

\subsection{Ablation Study}\label{sec:ablation}

Table~\ref{tab:ablation} decomposes CQTR on RT-DETR-R50 into counterfactual scaling, trajectory recalibration, conflict suppression, and stream routing. Counterfactual scaling alone contributes most of the gain, reaching 54.18, 12.99, and 17.13 AP on COCO, TinyPerson, and VisDrone, respectively. Under overall AP, the trajectory rule performs best on COCO but is slightly below scale-only inference on TinyPerson and VisDrone. Conflict suppression performs better on the two cross-domain datasets but slightly decreases COCO AP. Using unlabeled statistics to select the corresponding branch, stream routing improves scale-only inference by 0.03, 0.05, and 0.10 AP, respectively. These results explain why the later components should not be applied unconditionally. They are weak ranking corrections for different internal failure modes, rather than additional scale enhancers.

\subsection{Qualitative Results}\label{sec:qualitative}

As shown in Fig.~\ref{fig:qualitative}, CQTR recovers small objects missed by the frozen baseline on all three datasets under a shared confidence threshold. The numbers of true positives in the COCO, TinyPerson, and VisDrone examples increase from 18, 9, and 43 to 23, 28, and 61, while false negatives decrease from 18, 68, and 131 to 13, 49, and 113. False positives also decrease on COCO and TinyPerson, while VisDrone incurs only one additional false positive. CQTR therefore improves small-object recall in crowded scenes and complex backgrounds while maintaining prediction reliability. The corrections motivated by internal interpretation ultimately translate into observable detection gains.

\subsection{Mechanistic Interpretability and Analysis}\label{sec:interpretability}

\begin{figure}[t]
\centering
\includegraphics[width=\linewidth]{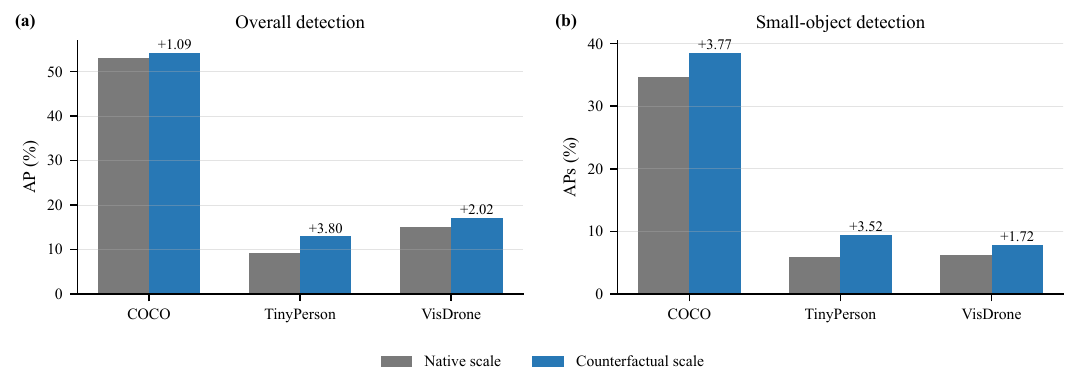}
\vspace{-8mm}
\caption{Activation of latent scale knowledge. AP and $\mathrm{AP}_\mathrm{s}$ are compared between the frozen baseline and the counterfactual scale branch alone.}
\label{fig:latent-scale}
\end{figure}

\textbf{Activation of latent scale knowledge.} Fig.~\ref{fig:latent-scale} compares RT-DETR-R50 under native inference and the counterfactual scale branch alone. Without updating any parameter, scale intervention improves both AP and $\mathrm{AP}_\mathrm{s}$ on COCO, TinyPerson, and VisDrone. 
The $\mathrm{AP}_\mathrm{s}$ gain is especially pronounced on COCO. These consistent improvements show that the relevant capability can be elicited from frozen parameters rather than necessarily relearned through target-domain training. The result supports our hypothesis that frozen detectors contain scale-sensitive representations that standard inference does not fully activate.

\textbf{Reliability of query-trajectory evidence.} We next test whether spatial-semantic trajectories explain candidate reliability. Using a fixed random seed, we sample 100 images from each evaluation set. Candidates with an area no greater than ${32}^{2}$ and a confidence of at least 0.05 are divided into five equal-frequency bins according to trajectory evidence. 
\begin{figure}[t]
\centering
\includegraphics[width=\linewidth]{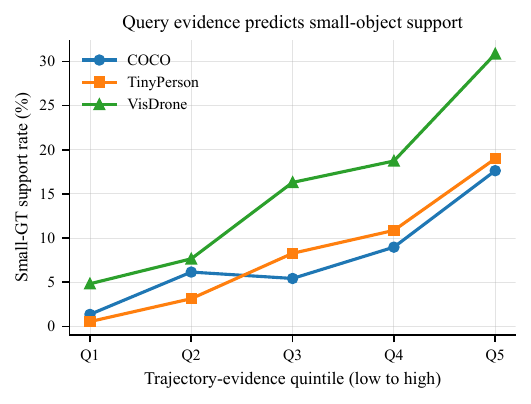}
\vspace{-8mm}
\caption{Query-trajectory evidence versus small-object ground-truth support. Q1 and Q5 denote the lowest and highest trajectory-evidence quintiles.}
\vspace{-4mm}
\label{fig:trajectory-evidence}
\end{figure}
A candidate receives ground-truth support when its IoU with a small ground-truth object of the same class is at least 0.5. These annotations are used only for post hoc analysis and do not participate in CQTR inference or routing. As shown in Fig.~\ref{fig:trajectory-evidence}, the small-object support rate generally rises with trajectory evidence. From the lowest to the highest evidence bin, the support rate increases from 1.35\% to 17.62\% on COCO, from 0.54\% to 18.98\% on TinyPerson, and from 4.82\% to 30.84\% on VisDrone. Trajectory evidence is therefore not an arbitrary score adjustment. It is directly associated with the correctness of small-object candidates.

\begin{figure}[t]
\centering
\includegraphics[width=\linewidth]{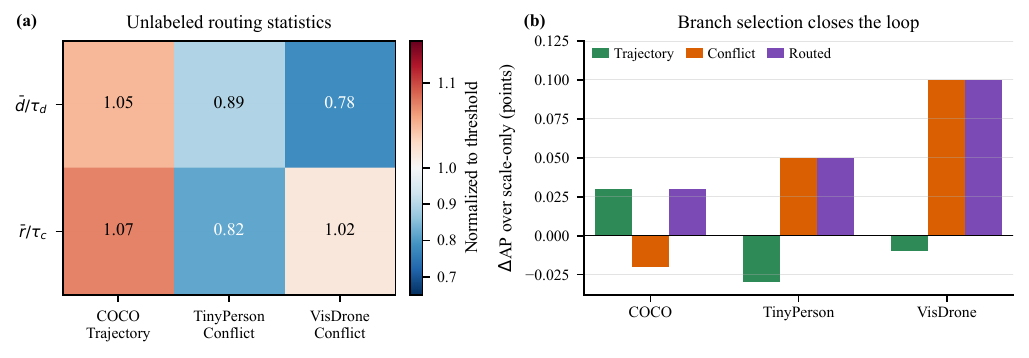}
\vspace{-8mm}
\caption{Closed loop of unlabeled stream routing. Left: routing statistics normalized by their thresholds. Right: AP changes of the trajectory, conflict, and routed branches relative to the scale-only branch.}
\label{fig:routing}
\end{figure}

\textbf{Closed loop of unlabeled stream routing.} Fig.~\ref{fig:routing} tests whether unlabeled stream statistics identify the more effective interpretation mechanism. The left panel reports normalized trajectory-evidence dispersion $\bar d/{\tau }_{d}$ and average maximum conflict strength $\bar r/{\tau }_{c}$, computed from an independent unlabeled training subset. The right panel shows the AP change of each correction branch relative to the scale-only branch. Both COCO statistics exceed their thresholds, so the router selects trajectory recalibration. TinyPerson and VisDrone are routed to conflict suppression. For these three RT-DETR-R50 streams, every routing decision agrees with the branch that performs better under full evaluation. Internal statistics derived from unlabeled images can therefore predict the effective correction mechanism, closing the loop from internal state to routing decision and performance improvement without selecting a branch post hoc from test AP.

\textbf{Decoder-internal evidence. }We further visualize semantic query density and spatial trajectories across the final four decoder layers. High-evidence queries remain concentrated around small objects and converge more stably than low-evidence queries, providing direct internal support for trajectory-based reliability estimation. Detailed definitions, statistics, and visualizations are provided in Appendix~\ref{app:trajectory_visualization}.

\section{Conclusion}\label{sec:conclusion}

We presented CQTR, a training-free framework for improving small-object detection with frozen detectors. Through a counterfactual scale branch, spatial-semantic query-trajectory interpretation, and unlabeled reliability routing, CQTR activates latent scale knowledge and suppresses cross-scale conflicts. Experiments across multiple detectors and three datasets demonstrate its effectiveness, generality, and interpretability. Future work will investigate lower-cost single-pass interventions and extensions to open-world detection.

\bibliographystyle{IEEEtran}
\bibliography{ref}

\appendix
\section{Appendix}
\subsection{Full Quantitative Results}
\label{app:full_results}
Figures~\ref{fig:main_ap} and~\ref{fig:main_aps} in the main paper visualize the overall AP and small-object AP trends across the three datasets. To provide complete quantitative evidence,
Tables~\ref{tab:coco}--\ref{tab:visdrone} report AP, $\mathrm{AP}_{50}$, $\mathrm{AP}_{75}$, and $\mathrm{AP}_{s}$ for all 27 detector-dataset combinations. Values in parentheses denote absolute improvements over the corresponding frozen baselines. All results are obtained without updating the model parameters or using target-domain training annotations.

\textbf{COCO results.} Table~\ref{tab:coco} provides the complete results on COCO val2017. CQTR improves all four evaluation metrics for every frozen detector. The highest AP of 54.32 is obtained by RT-DETRv2-R50 with
CQTR. RT-DETR-R50 exhibits the largest $\mathrm{AP}_\mathrm{s}$ improvement, increasing from 34.74 to 38.55. These detailed results confirm the trend in Figures~\ref{fig:main_ap} and and~\ref{fig:main_aps}, where the improvement in small-object detection is more pronounced than the overall AP gain.

\textbf{Cross-domain TinyPerson results.} Table~\ref{tab:tinyperson} reports the complete cross-domain
results on TinyPerson. CQTR consistently improves AP, $\mathrm{AP}_{50}$, $\mathrm{AP}_{75}$, and $\mathrm{AP}_\mathrm{s}$ across all nine detectors. RT-DETR-R50 achieves the largest improvement, with AP increasing from 9.19 to 13.04 and $\mathrm{AP}_\mathrm{s}$ increasing from 5.89 to 9.60. The consistent gains across different detector families demonstrate that CQTR transfers scale-sensitive knowledge beyond the COCO pretraining distribution.

\textbf{Cross-domain VisDrone2019 results.} As reported in Table~\ref{tab:visdrone}, CQTR also improves every
detector on VisDrone2019-DET val. DEIM-D-FINE-S obtains the largest AP gain, increasing from 14.50 to 17.16, together with an
$\mathrm{AP}_\mathrm{s}$ improvement from 6.22 to 8.61. RT-DETRv2-R50 achieves the highest absolute AP of 17.49. These results support the effectiveness of CQTR in dense aerial scenes with severe scale variation and background interference.

\begin{figure*}[t]
\centering
\includegraphics[width=\linewidth]{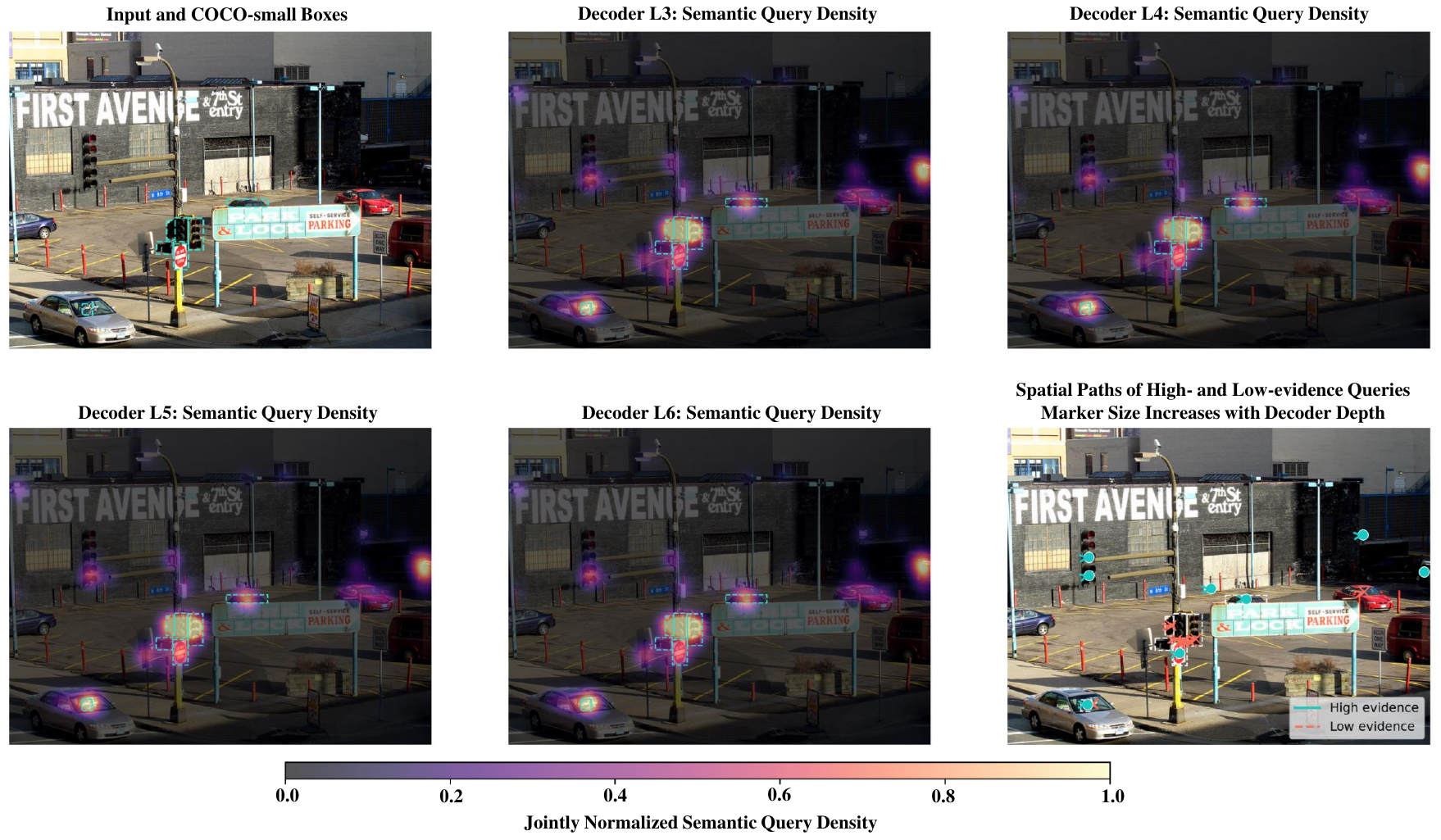}
\vspace{-8mm}
\caption{Decoder-internal spatial-semantic query trajectories. Semantic-density maps and representative trajectories from the final four decoder layers show that high-evidence queries converge more stably around small objects than low-evidence queries, with ground truth used only for post hoc visualization.}
\vspace{-4mm}
\label{fig:query-trajectories}
\end{figure*}

\subsection{Visualization of Decoder Query Trajectories}
\label{app:trajectory_visualization}
To further explain the internal meaning of spatial-semantic trajectory evidence, we project query states from the final four decoder layers onto the original image. Let ${\mu }_{q}^{\left(l\right)}$ be the center of the box predicted by query $q$ at layer $l$, and let ${p}_{q}^{\left(l\right)}\left({y}_{q}\right)$ denote the response at that layer to its final class ${y}_{q}$. The semantic query density at layer $l$ is defined as
\begin{equation}
H^{(l)}(u,v)=\sum_{q\in Q_s}p_q^{(l)}(y_q)\exp\!\left(-\frac{\lVert(u,v)-\mu_q^{(l)}\rVert_2^2}{2\sigma^2}\right),
\end{equation}
where ${Q}_{s}$ is the set of queries whose final predicted area does not exceed the small-object threshold. All layers use a shared normalization range, allowing their colors to be compared directly.

As shown in Fig.~\ref{fig:query-trajectories}, decoder query responses concentrate on small-object regions such as traffic lights and vehicles, and maintain high object-to-background response ratios in later layers. The image contains 80 small-object candidate queries. Their response ratios between small-object and background regions are 24.58, 26.44, 26.14, and 26.06 across the four layers, showing persistent concentration around small objects during iterative decoding. The lower-right panel further compares the spatial paths of high- and low-evidence queries. High-evidence queries remain spatially convergent across the last four layers, whereas low-evidence queries move back and forth more noticeably in the dense traffic-light region. The mean trajectory length of high-evidence queries is only 0.013\% of the image diagonal. The corresponding value for low-evidence queries is 0.184\%, approximately 14.3 times larger. CQTR trajectory evidence thus separates stably convergent queries from spatially uncertain ones and provides observable internal support for subsequent reliability recalibration.

\begin{table*}[t]
\centering
\small
\caption{Comparison on COCO val2017.}
\label{tab:coco}
\resizebox{\textwidth}{!}{%
\begin{tabular}{lcccc}
\toprule
Method & AP & $\mathrm{AP}_{50}$ & $\mathrm{AP}_{75}$ & $\mathrm{AP}_\mathrm{s}$ \\
\midrule
DETR-R50 & 36.51 & 55.83 & 37.74 & 13.65 \\
\rowcolor{blue!6} DETR-R50 + CQTR & 38.16 \gain{+1.65} & 57.51 \gain{+1.68} & 40.17 \gain{+2.43} & 15.77 \gain{+2.12} \\
\addlinespace[1pt]
Deformable-DETR-R50 & 40.45 & 58.58 & 43.83 & 19.44 \\
\rowcolor{blue!6} Deformable-DETR-R50 + CQTR & 41.98 \gain{+1.53} & 60.26 \gain{+1.68} & 45.78 \gain{+1.95} & 22.08 \gain{+2.64} \\
\addlinespace[1pt]
DAB-DETR-R50 & 36.35 & 56.40 & 37.64 & 13.61 \\
\rowcolor{blue!6} DAB-DETR-R50 + CQTR & 37.94 \gain{+1.59} & 57.72 \gain{+1.32} & 39.97 \gain{+2.33} & 15.74 \gain{+2.13} \\
\addlinespace[1pt]
DINO-R50 & 44.62 & 61.70 & 48.26 & 24.03 \\
\rowcolor{blue!6} DINO-R50 + CQTR & 46.23 \gain{+1.61} & 63.32 \gain{+1.62} & 50.33 \gain{+2.07} & 26.51 \gain{+2.48} \\
\addlinespace[1pt]
RT-DETR-R50 & 53.09 & 71.23 & 57.73 & 34.74 \\
\rowcolor{blue!6} RT-DETR-R50 + CQTR & 54.21 \gain{+1.12} & 71.86 \gain{+0.63} & 59.19 \gain{+1.46} & 38.55 \gain{+3.81} \\
\addlinespace[1pt]
RT-DETRv2-R50 & 53.40 & 71.56 & 57.48 & 36.06 \\
\rowcolor{blue!6} RT-DETRv2-R50 + CQTR & 54.32 \gain{+0.92} & 72.22 \gain{+0.66} & 58.88 \gain{+1.40} & 38.01 \gain{+1.95} \\
\addlinespace[1pt]
RT-DETRv3-R50 & 52.72 & 70.95 & 57.26 & 35.15 \\
\rowcolor{blue!6} RT-DETRv3-R50 + CQTR & 53.72 \gain{+1.00} & 71.55 \gain{+0.60} & 58.68 \gain{+1.42} & 38.15 \gain{+3.00} \\
\addlinespace[1pt]
D-FINE-S & 48.49 & 65.42 & 52.58 & 29.30 \\
\rowcolor{blue!6} D-FINE-S + CQTR & 49.58 \gain{+1.09} & 66.48 \gain{+1.06} & 53.94 \gain{+1.36} & 31.70 \gain{+2.40} \\
\addlinespace[1pt]
DEIM-D-FINE-S & 48.96 & 65.88 & 53.04 & 30.38 \\
\rowcolor{blue!6} DEIM-D-FINE-S + CQTR & 49.71 \gain{+0.75} & 66.69 \gain{+0.81} & 54.23 \gain{+1.19} & 32.57 \gain{+2.19} \\
\bottomrule
\end{tabular}%
}
\vspace{-3mm}
\end{table*}

\begin{table*}[t]
\centering
\small
\caption{Cross-domain COCO-style comparison on TinyPerson test.}
\label{tab:tinyperson}
\resizebox{\textwidth}{!}{%
\begin{tabular}{lcccc}
\toprule
Method & AP & $\mathrm{AP}_{50}$ & $\mathrm{AP}_{75}$ & $\mathrm{AP}_\mathrm{s}$ \\
\midrule
DETR-R50 & 1.87 & 5.39 & 1.12 & 0.69 \\
\rowcolor{blue!6} DETR-R50 + CQTR & 2.72 \gain{+0.85} & 8.02 \gain{+2.63} & 1.22 \gain{+0.10} & 1.26 \gain{+0.57} \\
\addlinespace[1pt]
Deformable-DETR-R50 & 2.92 & 9.22 & 1.13 & 1.09 \\
\rowcolor{blue!6} Deformable-DETR-R50 + CQTR & 4.45 \gain{+1.53} & 13.37 \gain{+4.15} & 1.89 \gain{+0.76} & 2.02 \gain{+0.93} \\
\addlinespace[1pt]
DAB-DETR-R50 & 1.78 & 5.34 & 1.11 & 0.67 \\
\rowcolor{blue!6} DAB-DETR-R50 + CQTR & 3.01 \gain{+1.23} & 9.72 \gain{+4.38} & 1.35 \gain{+0.24} & 1.11 \gain{+0.44} \\
\addlinespace[1pt]
DINO-R50 & 4.28 & 13.56 & 1.98 & 2.11 \\
\rowcolor{blue!6} DINO-R50 + CQTR & 6.71 \gain{+2.43} & 20.50 \gain{+6.94} & 3.05 \gain{+1.07} & 3.89 \gain{+1.78} \\
\addlinespace[1pt]
RT-DETR-R50 & 9.19 & 26.24 & 4.63 & 5.89 \\
\rowcolor{blue!6} RT-DETR-R50 + CQTR & 13.04 \gain{+3.85} & 34.83 \gain{+8.59} & 7.11 \gain{+2.48} & 9.60 \gain{+3.71} \\
\addlinespace[1pt]
RT-DETRv2-R50 & 9.45 & 26.16 & 4.82 & 5.97 \\
\rowcolor{blue!6} RT-DETRv2-R50 + CQTR & 11.66 \gain{+2.21} & 32.23 \gain{+6.07} & 5.96 \gain{+1.14} & 7.88 \gain{+1.91} \\
\addlinespace[1pt]
RT-DETRv3-R50 & 9.12 & 25.37 & 4.74 & 5.80 \\
\rowcolor{blue!6} RT-DETRv3-R50 + CQTR & 12.82 \gain{+3.70} & 33.98 \gain{+8.61} & 7.28 \gain{+2.54} & 9.15 \gain{+3.35} \\
\addlinespace[1pt]
D-FINE-S & 7.24 & 20.74 & 3.67 & 4.37 \\
\rowcolor{blue!6} D-FINE-S + CQTR & 9.80 \gain{+2.56} & 27.39 \gain{+6.65} & 4.99 \gain{+1.32} & 6.77 \gain{+2.40} \\
\addlinespace[1pt]
DEIM-D-FINE-S & 7.80 & 21.88 & 4.15 & 4.96 \\
\rowcolor{blue!6} DEIM-D-FINE-S + CQTR & 9.80 \gain{+2.00} & 27.29 \gain{+5.41} & 5.39 \gain{+1.24} & 6.73 \gain{+1.77} \\
\bottomrule
\end{tabular}%
}
\vspace{-3mm}
\end{table*}

\begin{table*}[t]
\centering
\small
\caption{Cross-domain COCO-style comparison on VisDrone2019-DET val.}
\label{tab:visdrone}
\resizebox{\textwidth}{!}{
\begin{tabular}{lcccc}
\toprule
Method & AP & $\mathrm{AP}_{50}$ & $\mathrm{AP}_{75}$ & $\mathrm{AP}_\mathrm{s}$ \\
\midrule
DETR-R50 & 4.31 & 8.95 & 3.65 & 0.60 \\
\rowcolor{blue!6} DETR-R50 + CQTR & 5.12 \gain{+0.81} & 10.86 \gain{+1.91} & 4.20 \gain{+0.55} & 0.96 \gain{+0.36} \\
\addlinespace[1pt]
Deformable-DETR-R50 & 5.74 & 11.75 & 4.88 & 1.34 \\
\rowcolor{blue!6} Deformable-DETR-R50 + CQTR & 6.99 \gain{+1.25} & 14.17 \gain{+2.42} & 6.13 \gain{+1.25} & 2.04 \gain{+0.70} \\
\addlinespace[1pt]
DAB-DETR-R50 & 4.59 & 9.54 & 4.02 & 0.68 \\
\rowcolor{blue!6} DAB-DETR-R50 + CQTR & 6.14 \gain{+1.55} & 12.73 \gain{+3.19} & 5.23 \gain{+1.21} & 1.11 \gain{+0.43} \\
\addlinespace[1pt]
DINO-R50 & 7.86 & 15.53 & 7.03 & 2.36 \\
\rowcolor{blue!6} DINO-R50 + CQTR & 9.25 \gain{+1.39} & 18.14 \gain{+2.61} & 8.33 \gain{+1.30} & 3.65 \gain{+1.29} \\
\addlinespace[1pt]
RT-DETR-R50 & 15.11 & 27.11 & 14.57 & 6.16 \\
\rowcolor{blue!6} RT-DETR-R50 + CQTR & 17.23 \gain{+2.12} & 30.80 \gain{+3.69} & 16.46 \gain{+1.89} & 7.85 \gain{+1.69} \\
\addlinespace[1pt]
RT-DETRv2-R50 & 15.06 & 27.16 & 14.43 & 6.02 \\
\rowcolor{blue!6} RT-DETRv2-R50 + CQTR & 17.49 \gain{+2.43} & 30.95 \gain{+3.79} & 16.85 \gain{+2.42} & 8.10 \gain{+2.08} \\
\addlinespace[1pt]
RT-DETRv3-R50 & 15.07 & 26.83 & 14.59 & 6.22 \\
\rowcolor{blue!6} RT-DETRv3-R50 + CQTR & 17.27 \gain{+2.20} & 30.81 \gain{+3.98} & 16.68 \gain{+2.09} & 7.92 \gain{+1.70} \\
\addlinespace[1pt]
D-FINE-S & 12.98 & 23.55 & 12.44 & 5.55 \\
\rowcolor{blue!6} D-FINE-S + CQTR & 15.27 \gain{+2.29} & 26.90 \gain{+3.35} & 14.87 \gain{+2.43} & 7.53 \gain{+1.98} \\
\addlinespace[1pt]
DEIM-D-FINE-S & 14.50 & 26.24 & 13.68 & 6.22 \\
\rowcolor{blue!6} DEIM-D-FINE-S + CQTR & 17.16 \gain{+2.66} & 30.05 \gain{+3.81} & 16.84 \gain{+3.16} & 8.61 \gain{+2.39} \\
\bottomrule
\end{tabular}
}
\vspace{-3mm}
\end{table*}

\end{document}